%% file: MatrixZero.tex
\documentclass{article} 
\usepackage[final]{neurips_2025} 

\usepackage[utf8]{inputenc} % allow utf-8 input
\usepackage[T1]{fontenc} % use 8-bit T1 fonts
\usepackage{hyperref} % hyperlink
\usepackage{subcaption}
\hypersetup{
 colorlinks=true,
 linkcolor=red,
 citecolor=cyan,
 filecolor=magenta, 
 urlcolor=magenta,
 }
\usepackage{url} % simple URL typesetting
\usepackage{booktabs} % professional-quality tables
\usepackage{amsfonts} % blackboard math symbols
\usepackage{nicefrac} % compact symbols for 1/2, etc.
\usepackage{microtype} % microtypography
\usepackage{xcolor} % colors
\usepackage{amsmath} 
\usepackage{xspace} 
\usepackage{enumitem} 
\usepackage{textcomp}
\usepackage{stfloats}
\usepackage{verbatim}
\usepackage{wrapfig}
\usepackage{graphicx}
\usepackage{subcaption}
\usepackage{amssymb}
\usepackage{cite}
\usepackage{tikz}
\usepackage{algorithm}
\usepackage{algpseudocode}

\usepackage{multicol,multirow} 
\usepackage{threeparttable} 
\usepackage{pgfplots}
\pgfplotsset{compat=1.18}  % 兼容设置，1.18 是常见版本

\title{Matrix-game 2.0: An open-source, real-time, and streaming interactive world model}

\author{%
 \textbf{Xianglong He}\thanks{Equal contribution.}\quad
 \textbf{Chunli Peng}$^{*}$\quad
 \textbf{Zexiang Liu}$^{*}$\quad
 \textbf{Boyang Wang}$^{*}$\thanks{Project Lead.}\quad
\textbf{Yifan Zhang}\quad
\textbf{Qi Cui}\quad \\
 \textbf{Fei Kang}\quad
\textbf{Biao Jiang}\quad
\textbf{Mengyin An}\quad
\textbf{Yangyang Ren}\quad
\textbf{Baixin Xu}\quad
\textbf{Hao-Xiang Guo}\quad\\
\textbf{Kaixiong Gong}\quad
\textbf{Size Wu}\quad
\textbf{Wei Li}\quad
\textbf{Xuchen Song}\quad
\textbf{Yang Liu}$^{\dag\ddag}$\quad
\textbf{Yangguang Li}$^{\dag}$\thanks{Correspondence: liyangguang256@gmail.com}\quad
 \textbf{Yahui Zhou} \\
 \\
 Skywork AI \\
 Project page: \href{https://matrix-game-v2.github.io/}{Matrix-Game-2.0-Homepage}
}

\begin{document}
\maketitle

\begin{figure}[htbp]
  \centering
  % \hspace{-1cm}
  % \vspace{-1cm}
  \includegraphics[width=\textwidth,page=1]{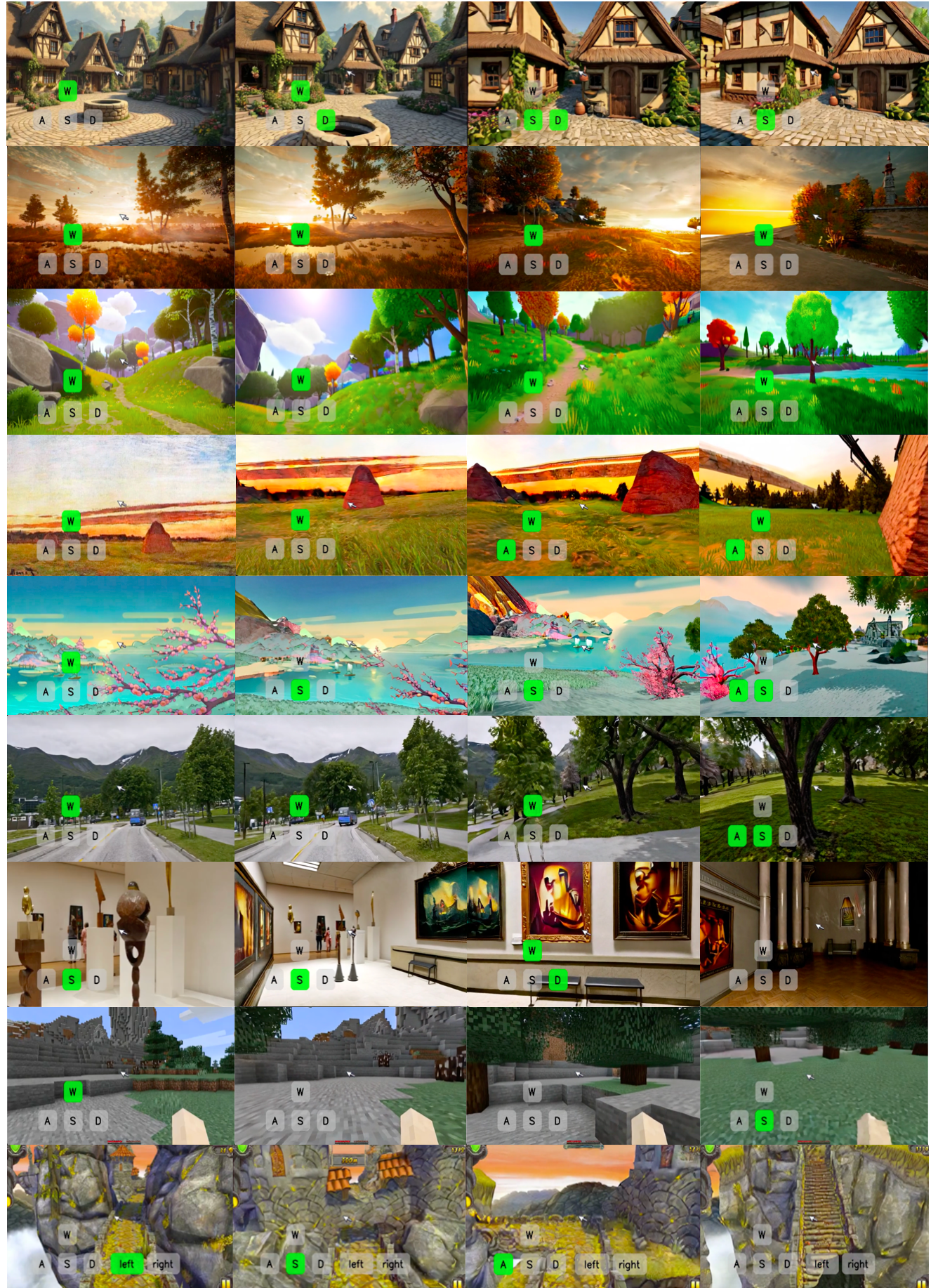}  % page=1 插入第1页
  \caption{\textbf{Real-time Interactive Generation Results.} We introduce Matrix-Game 2.0, a real-time interactive video generation model. By integrating action modules and few-step distillation, it can auto-regressively produce high-quality interactive videos given an input image in 25 FPS. The demonstrated results cover various scenes and diverse styles, demonstrating its powerful generation capabilities.}
  \label{fig:overal_samples}
\end{figure}

 \begin{figure}[htbp]
  \centering
  \includegraphics[width=\textwidth]{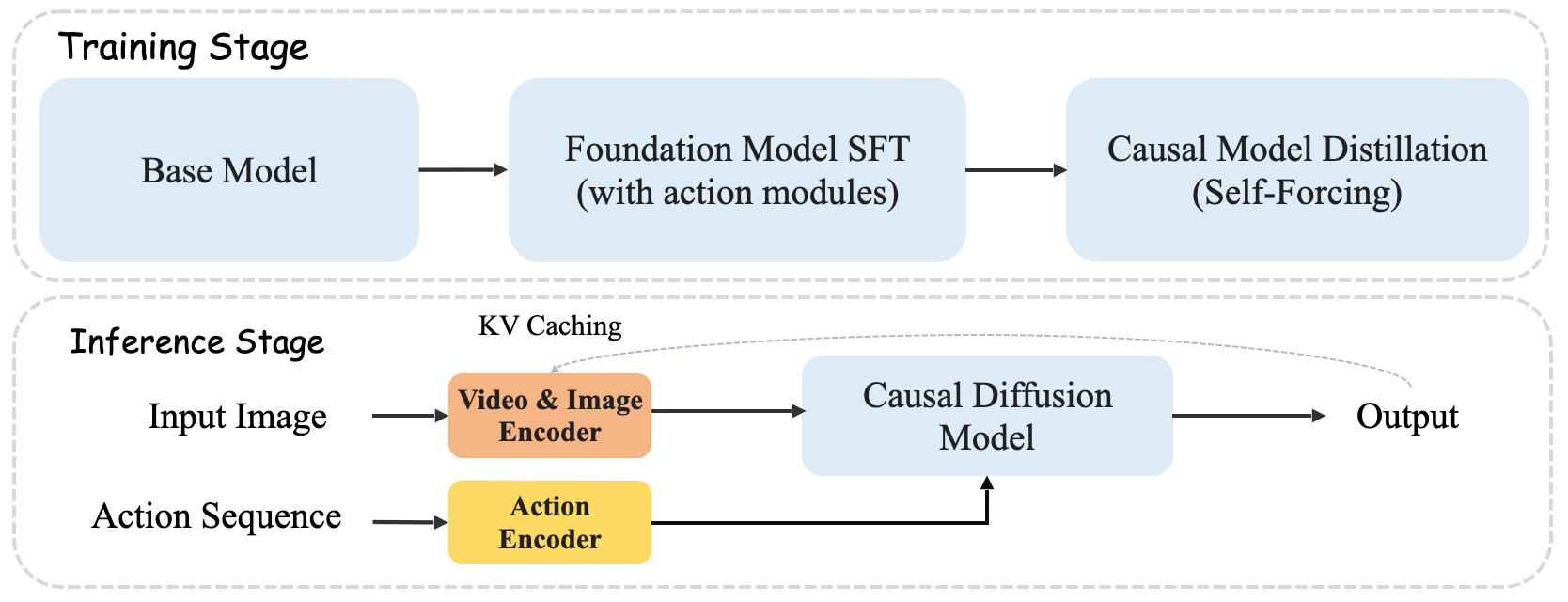}
  \vspace{-0.4cm}
  \caption{\textbf{Pipelines of Matrix-Game 2.0.}}
  \label{fig:overall_architecture}
  \vspace{-0.8cm}
\end{figure}
\vspace{-0.99cm}
% abstract
\begin{abstract}
Recent advances in interactive video generations have demonstrated diffusion model's potential as world models by capturing complex physical dynamics and interactive behaviors. However, existing interactive world models depend on bidirectional attention and lengthy inference steps, severely limiting real-time performance. Consequently, they are hard to simulate real-world dynamics, where outcomes must update instantaneously based on historical context and current actions. To address this, we present Matrix-Game 2.0, an interactive world model generates long videos on-the-fly via few-step auto-regressive diffusion. Our framework consists of three key components: (1) A scalable data production pipeline for Unreal Engine and gaming environments to effectively produce massive amounts ($\sim$ 1200 hours) of video data with diverse interaction annotations; (2) An action injection module that enables frame-level mouse and keyboard inputs as interactive conditions; (3) A few-step distillation based on the casual architecture for real-time and streaming video generation. Matrix-Game 2.0 can generate high-quality minute-level videos across diverse scenes at an ultra-fast speed of 25 FPS. We open-source our model weights and codebase to advance research in interactive world modeling.
% , and data production pipeline
\end{abstract}

\section{Introduction} 
\input{sec/introduction}

% related work
\section{Related Work}
\input{sec/related_work}

% \clearpage

% data Pipeline
\section{Data Pipeline Development}
\label{sec:data_pipeline}
\input{sec/Data_Pipeline}

% method
\section{Methods}
\input{sec/summary}

% - foundation model archtecture
\subsection{Foundation Model Architecture}\label{sec:foundtion_model_architecture}
\input{sec/foundation_model}

% - Long video generation
\subsection{Real-time Interactive Auto-Regressive Video Generation}\label{sec:self-forcing}
\input{sec/self_forcing}

% experiment
\section{Experiments}

\input{sec/experiments}

% conclusion
\section{Conclusion} 
\input{sec/conclusion}
% references
\clearpage
{ 
\small
\bibliography{neurips_2025}
\bibliographystyle{plain} 
}

\end{document}

%% file: sec/introduction.tex
World models~\cite{genie3,po2025long,li2025hunyuan} have gained significant attention due to their capability to understand real-world interactions and predict future states~\cite{yang2023learning}. By enabling intelligent agents to perceive their surroundings and respond to actions, these models reduce the cost of real-world trials and facilitate interactive simulation. Consequently, world models show great promise in fields such as game engines~\cite{valevski2024diffusion,chen2025deepverse,oasis2024}, autonomous driving~\cite{hu2023gaia}, and spatial intelligence~\cite{bardes2023v, worldlabs2025generating, parkerholder2024genie2, genie3}.

Recent advances in video generation models~\cite{lin2025autoregressive, blattmann2023stable,kong2024hunyuanvideo,wan2025wanopenadvancedlargescale,teng2025magi} have shown remarkable progress in learning knowledge from large-scale real-world datasets, ranging from physical laws to interactive scenes. This demonstrates their huge potential to serve as world models. Among the various research directions in this domain, interactive long video generation~\cite{che2024gamegen,bruce2024genie} has become increasingly important due to its practical applications, where long videos should be generated in real-time in response to a continuous stream of user input. Specifically, when conditioned on user actions like camera movements and keyboard inputs, the model generates frames causally, enabling real-time user interaction.

Despite impressive progress in interactive video generation, existing methods suffer from several significant challenges:

$\bullet$ Lack of large-scale, high-quality interactive video datasets with rich annotations for training, such as accurate actions and camera dynamics, due to the high cost and difficulty of collection.

$\bullet$ Latency issues with bidirectional video diffusion models~\cite{zhang2025matrixgame,li2025hunyuan,mao2025yume}, where generating a single frame requires processing the entire video. This makes them unsuitable for real-time, streaming applications where the model must adapt to dynamic user commands and produce frames on the fly. The quadratic scaling of compute and memory requirements with respect to frame length, along with the high number of denoising iterations, makes long video generation computationally intensive and economically impractical.

$\bullet$ Severe error accumulation in existing auto-regressive video diffusion models~\cite{jin2024pyramidal,teng2025magi,chen2025deepverse}. While these models generate the next frame based on previous frames, they often suffer from error accumulation during generation, leading to degraded video quality over time.

To address these critical challenges in real-time interactive generation, we present Matrix-Game 2.0 - a novel framework specifically designed to achieve both real-time performance and robust generalization across diverse scenarios. First, our technical core features a video diffusion transformer with integrated action control modules, distilled into a causal few-step auto-regressive model via Self-Forcing~\cite{huang2025self} based techniques. This architecture supports both training and inference through an efficient KV caching mechanism, achieving 25 FPS generation on a single H100 GPU while maintaining minute-long temporal consistency and precise action controllability - even in complex wild scenes beyond the training distribution.

The model's strong generalization capability is enabled by another innovation of ours: a comprehensive data production pipeline that solves fundamental limitations in interactive training data. The pipeline is based on Unreal Engine, including a Navigation Mesh-based Path Planning System for data diversity and Quaternion Precision Optimization modules for accurate camera control. Moreover, for gaming environment, we developed a data recording system using Script Hook integration, which enables synchronized capture of visual content with corresponding user interactions. Together, these components produce large-scale datasets with frame-level annotations, addressing two critical needs: (1) precise alignment between visual content and control signals, and (2) effective modeling of dynamic in-game interactions.

By simultaneously tackling the challenges of efficiency and controllability, Matrix-Game 2.0 makes significant strides in world modeling by introducing an efficient framework tailored for real-time simulation and interaction. To support continued progress in this area, we will release the code and weights of our pre-trained models.

%% file: sec/related_work.tex
\subsection{Controllable Video Generation}
With the rapid advancement of diffusion models~\cite{sohl2015deep,ho2020denoising,songscore}, significant progress has been made in visual content generation for videos~\cite{tian2024emo,zheng2024memo,zhang2023expanding,blattmann2023stable,lin2024open,yang2024cogvideox,HaCohen2024LTXVideo}. Most recent approaches have transitioned to bidirectional diffusion transformers~\cite{peebles2023scalable} and auto-regressive models~\cite{jin2024pyramidal,chen2025deepverse}, enabling modern video diffusion models to synthesize high-quality, temporally coherent, and substantially longer videos. This rapid evolution of video generation~\cite{openai2024worldsim,parkerholder2024genie2,agarwal2025cosmos,yanglearning} has further driven the development of world models that leverage video diffusion techniques~\cite{chen2024diffusion,yin2025slow,huang2025self} to implicitly learn physical laws, object dynamics, and causality for complex environment simulation. 

Controllable video generation serves as a core component of world simulation. Generally, control signals span multiple modalities and can be categorized into scene controllability and action controllability. Extensive prior work has explored scene controllability, including~\cite{ren2025gen3c, kong2024hunyuanvideo, wang2025wan, HaCohen2024LTXVideo}, which leverages text, images, or 3D scene priors to regulate the scenes in generated videos. Beyond scene control, action controllability, achieved through camera angles~\cite{yang2024direct} or trajectories~\cite{he2024cameractrl,wang2024motionctrl}, has also emerged as a prominent research focus. These efforts have yielded promising advancements in both the visual quality and controllability of generated videos. Certain world model-based approaches~\cite{oasis2024, parkerholder2024genie2, yu2025gamefactory, guo2025mineworld, zhang2025matrixgame} further support both scene and action controllability. However, constrained by computational resources and video length limitations, most existing models still struggle to achieve real-time video generation.

\subsection{Long-context Video Generation}
Current video generation models are typically constrained to videos of $\leq$ 5 seconds due to limited long video training data and prohibitive computational costs. Existing methods for long-context video generation can be broadly categorized into two types: Combination of multiple short video segments, and auto-regressive generation approaches. For multi-segment video generation, a simple yet effective approach is to generate multiple overlapping segments of fixed length~\cite{chen2023seine, qiu2023freenoise,zhang2025matrixgame}. 

While some other works~\cite{yin2023nuwa, zhao2024moviedreamer} adopt a two-stage pipeline, first generating key frames and then applying frame interpolation. In contrast, auto-regressive models offer a natural advantage for variable-length video generation. For example, methods such as Diffusion Forcing~\cite{chen2024diffusion}, CausVid~\cite{yin2025slow}, and Self-Forcing~\cite{huang2025self} combine auto-regressive modeling with diffusion techniques to achieve promising results in long video synthesis. However, these approaches remain largely confined to conventional Text-to-Video (T2V) and Image-to-Video (I2V) tasks, leaving the challenge of generating long, interactive videos largely unexplored.

\subsection{Real-Time Video Generation}
Diverse approaches currently exist to achieve real-time video generation. The primary methods involve increasing the compression ratio of the VAE, performing knowledge distillation to reduce the number of sampling steps in diffusion models, or combining KV Cache with a causal transformer to autoregressively infer the next frame. LTX-Video~\cite{HaCohen2024LTXVideo} achieves generation times shorter than the video duration on an H100 GPU by optimizing VAE compression ratios and applying model distillation techniques~\cite{song2023consistency,yin2024one,zhang2023hipa}. Works such as Next-Frame Diffusion~\cite{cheng2025playing}, Self-Forcing~\cite{huang2025self}, CausVid~\cite{yin2025slow}, and Oasis~\cite{oasis2024} leverage the characteristics of autoregressive models, combined with knowledge distillation, to enable efficient few-step generation. Although these works achieve real-time video generation, most of them cannot support real-time interaction. While Oasis manages to enable real-time interaction, its visual quality degrades rapidly during the inference of long videos. In our work, we follow the training paradigm of Self-Forcing~\cite{huang2025self} to allow few-step inference, achieving not only fast long video generation but also maintaining stable and consistent frame quality.

%% file: sec/Data_Pipeline.tex
We design and implement comprehensive data production pipelines to facilitate large-scale training of Matrix-Game 2.0. Specifically, our work addresses two key challenges: (1) generating gaming video data precisely aligning with keyboard and camera signal annotations, and (2) enabling interactive video capture mechanisms powered by collision-aware navigation rules and reinforcement learning-trained agents to better model dynamic in-game interactions. For practical deployment, we develop and curate a diverse dataset production pipeline comprising both static and dynamic scenes sourced from the Unreal Engine and gaming simulation environment.

\begin{figure}[!htb]
 \centering
 \includegraphics[width=1\textwidth]{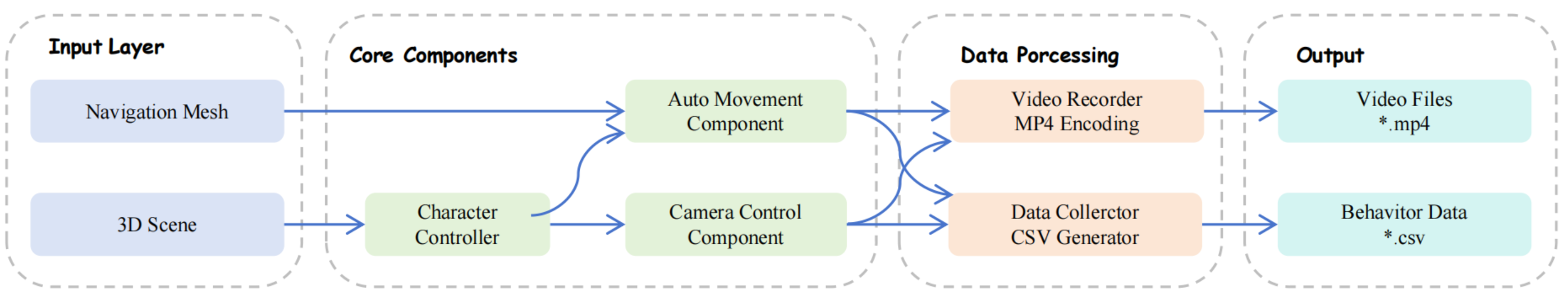}
 \caption{\textbf{Overview of Our Data Production Pipeline based on Unreal Engine.}}
 \label{fig:Unreal_pipeline}
 \vspace{-0.4cm}
\end{figure}

\subsection{Unreal Engine-based Data Production}
The development of high-performance interactive video generation models requires large-scale datasets featuring precisely synchronized visual content and control signals like precisely aligned keyboard input and camera parameters. While existing datasets often lack accurate temporal alignment between game-play footage and corresponding inputs, our Unreal Engine-based pipeline systematically addresses this gap through controlled synthetic data generation. Unreal Engine's precise environmental control and deterministic rendering make it particularly suitable for creating scalable, multi-modal training data with guaranteed annotation accuracy.

% \subsubsection{Framework Design}
As illustrated in Figure~\ref{fig:Unreal_pipeline}, our Unreal Engine-based data pipeline takes a navigation mesh and a 3D scene as input. The system then employs automated movement and camera control modules to simulate agent navigation and dynamic viewpoint transitions. Finally, the resulting visual data and corresponding action annotations are recorded and exported through an integrated MP4 encoder and CSV generator. 

The key innovations of our system comprise: (1) a navigation mesh-based path planning module to enable diverse trajectory generation; (2) a precise system input and camera control mechanism to ensure accurate action and viewpoint alignment; and (3) a structured post-processing pipeline for high-quality data curation. 
Detailed descriptions of each component are provided below.

\paragraph{Navigation Mesh-based Path Planning System.}
\begin{figure}[!htb]
 \centering
 \includegraphics[width=0.99\textwidth]{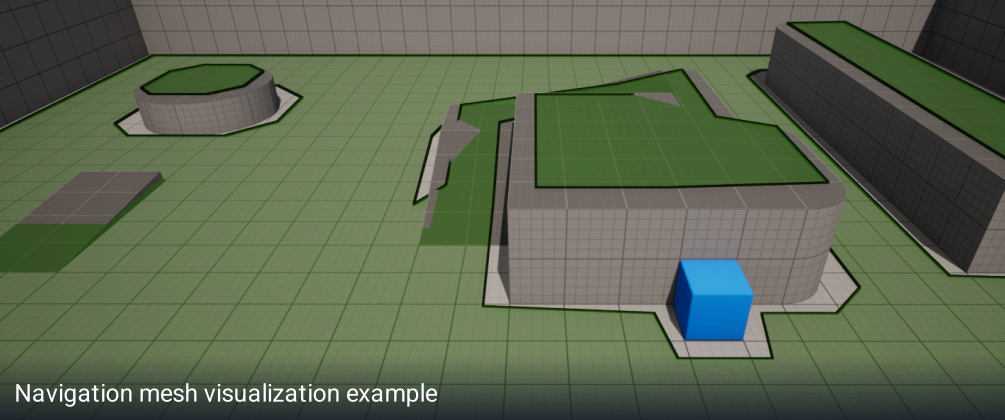}
 \caption{\textbf{An example for Our Navigation System.}}
 \label{fig:NavigationMesh}
\end{figure}
To enhance the realism and behavioral diversity of the generated training data, we developed an advanced navigation mesh–based path planning system that facilitates dynamic and adaptive movement of non-player characters (NPCs). This system supports real-time, deterministic path planning, a critical requirement for producing reproducible and high-fidelity training data.

Our implementation builds upon Unreal Engine's native NavMesh infrastructure, augmented with customized path-planning optimizations that reduce the average query latency to less than 2 ms. Furthermore, the system introduces controlled stochasticity in agent behavior, enabling diverse and contextually coherent movement patterns while strictly adhering to logical navigation constraints. This approach substantially enhances the richness of the training corpus by introducing realistic agent interaction dynamics and movement trajectories, thereby improving the generalization capacity of downstream video generation models. A navigation example is shown in Figure~\ref{fig:NavigationMesh}. The green area in the picture shows the area where the agent can move freely, preventing the agent from hitting the walls and getting stuck.

\paragraph{Reinforcement Learning-Enhanced Agent Training.}
To further improve the behavioral realism and decision-making capabilities of our data collection agents, we integrated a reinforcement learning (RL) framework alongside our collision-based navigation rules, which adopts typical RL methods like Proximal Policy Optimization (PPO)~\cite{schulman2017proximal}. The RL agents are trained with a reward function that combines collision avoidance, exploration efficiency, and trajectory diversity:

\begin{equation}
R_t = \alpha \cdot R_{collision} + \beta \cdot R_{exploration} + \gamma \cdot R_{diversity}
\end{equation}

where $R_{collision}$ penalizes collision events, $R_{exploration}$ rewards discovering new areas, and $R_{diversity}$ encourages diverse movement patterns. The collision-based rules serve as safety constraints during training, ensuring that RL agents maintain physical plausibility while learning optimal navigation strategies.

This hybrid approach combines the deterministic safety of rule-based collision avoidance with the adaptive intelligence of RL-trained behaviors, resulting in agents that can generate more realistic and diverse interaction patterns while maintaining data collection reliability.

\paragraph{Precise Input and Camera Control.}
We integrated Unreal Engine's Enhanced Input system to enable simultaneous capture of multiple keyboard inputs with millisecond-level precision. The system maintains a synchronized buffer of input events aligned with rendered frames to ensure accurate input–visual synchronization for training:

\begin{equation}
\text{Input}_{\text{frame}_i} = (\{k_1, k_2, ..., k_n\}, \text{timestamp}_i)
\end{equation}

where each input state $k_j$ represents a specific key press or release event aligned with frame $i$.

To eliminate a critical error rate of 0.2\% in camera rotation calculations, we implemented quaternion precision optimization by using double precision arithmetic in intermediate calculations.
This optimization reduced rotation errors to a level that is effectively negligible.

\begin{figure}[!htb]
 \centering
 \includegraphics[width=0.99\textwidth]{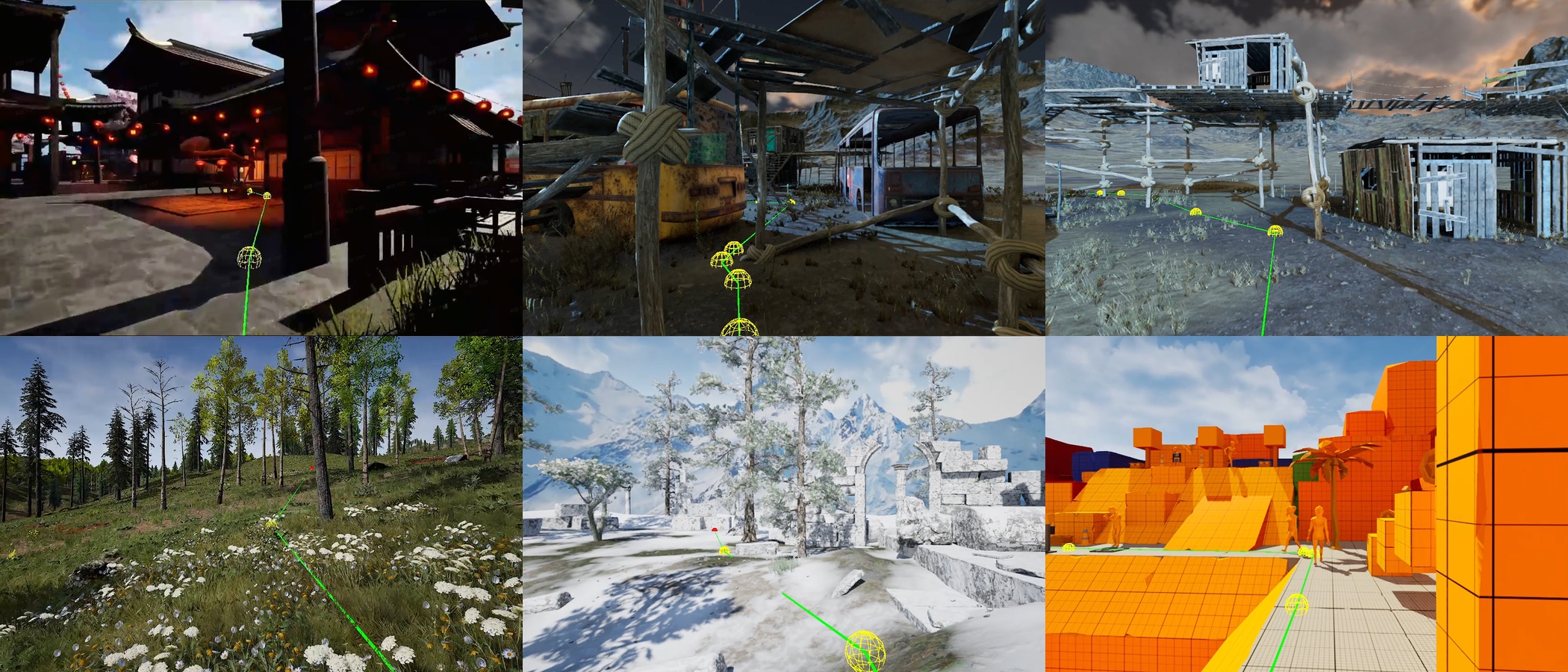}
 \caption{\textbf{Trajectory Examples of Collected Unreal Engine Data.}}
 \label{fig:UnrealTrajectory}
 \vspace{-0.2cm}
\end{figure}

\paragraph{Data Curation.}
We developed a video frame filtering algorithm based on OpenCV to detect and eliminate temporally redundant frames, thereby enhancing data efficiency.
A velocity-based validation mechanism was further introduced to identify and exclude invalid samples characterized by zero or negative velocity, which typically indicate stationary or physically implausible motion states:
\begin{equation}
\text{validity} = \begin{cases} 
1 & \text{if } ||\vec{v}|| > \epsilon \\
0 & \text{otherwise}
\end{cases}
\end{equation}

where $\vec{v}$ represents the velocity vector and $\epsilon$ is a small positive threshold to account for the precision of floating points. This criterion ensures the retention of only semantically meaningful motion data for subsequent model training. 

\paragraph{Multi-thread Pipeline Accelerating.}
The data processing pipeline was redesigned to support multi-thread execution, enabling dual-stream data production on a single RTX 3090 GPU. The system employs separate rendering threads in conjunction with shared memory pools for efficient resource utilization.
Some representative trajectory examples are illustrated in Figure~\ref{fig:UnrealTrajectory}. The green line segments represent the path of the agent. In complex scenarios, reasonable paths can also be planned.

\subsection{Gaming Interactive Data Recording System}

\begin{figure}[!htb]
 \centering
 \includegraphics[width=0.99\textwidth]{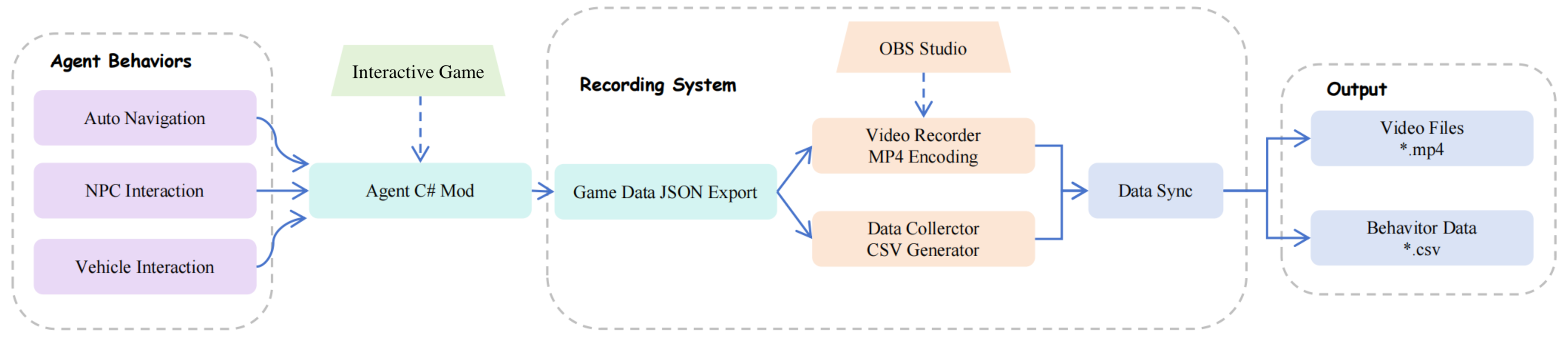}
 \caption{\textbf{Overview of Our Gaming Interactive Data Recording System.}}
 \label{fig:InteractiveDataArchitecture}
\end{figure}

To facilitate the acquisition of richly interactive dynamic scenes, we developed a comprehensive recording system within gaming using Script Hook integration, which enables synchronized capture of visual content with corresponding user actions. 

% \subsubsection{System Design}
We implemented a custom plugin architecture using Script Hook V to establish a recording pipeline within the gaming environment. The plugin simultaneously captures mouse and keyboard operations with frame-accurate synchronization. Each item collected includes the RGB frame and the corresponding mouse and keyboard operations. 

As illustrated in Figure~\ref{fig:InteractiveDataArchitecture}, Our system comprises three main components: Agent Behaviors, Interactive Game Environment, and Recording System. The Agent Behaviors module includes autonomous navigation, NPC interaction, and vehicle interaction capabilities, which are integrated into the interactive game through a custom C\# modification. The game exports behavioral data in JSON format to the Recording System, which utilizes OBS Studio for video capture with MP4 encoding and a Data Collector for CSV generation. A synchronization mechanism ensures temporal alignment between video frames and behavioral data, producing synchronized video files (.mp4) and behavioral datasets (.csv) as the final output. Dynamic control mechanisms, including autonomous navigation, NPC interaction, and vehicle interaction, can be selectively enabled to generate interactive scenarios from first-person or third-person perspectives. Environmental parameters such as vehicle density, NPC number, weather patterns, and time-of-day settings can be adjusted to simulate a wide variety of dynamic scenarios, enhancing the diversity and realism of the collected data. 
Specifically, the vehicle density parameter is configurable within the range $[0.1, 2.0]$ , while the NPC density parameter spans the interval $[0.2, 1.5]$.

To obtain an optimal viewpoint during vehicle navigation simulations, the system ensures precise camera alignment through per-tick positional updates, maintaining an optimal and consistent viewpoint relative to the vehicle throughout the simulation:

\begin{equation}
\text{Camera}_{position} = \text{Vehicle}_{position} + \text{offset} \times \text{rotation}
\end{equation}
Building upon the vehicle dynamics, the system infers and logs the corresponding keyboard inputs, thereby generating a comprehensive and temporally aligned interaction data encompassing velocity, acceleration and steering angle.

Additionally, we developed a runtime system to dynamically access navigation mesh information, facilitating intelligent camera positioning and motion prediction. This system performs queries on the navigation mesh data structure to extract spatial constraints and valid traversal paths, thereby enabling optimal planning of the camera trajectory.
The navigation mesh query process involves real-time spatial data retrieval and path validation to ensure that camera movements are confined within navigable regions while preserving optimal viewing angles for effective data acquisition.
\subsection{Quantative Data Evaluation}
We collected over 1.2 million video clips through our data curation pipeline, which demonstrated robust performance in several key metrics. The overall accuracy of the data exceeded 99\%, and the system achieved a 50-fold improvement in the precision of the camera rotation. Furthermore, the pipeline supported dual concurrent data streams per GPU, effectively doubling production efficiency.
For the interactive game environment which is complex and diverse, we can plan a reasonable path to prevent the agent from colliding or blocking, effectively improving the accuracy of the data.

%% file: sec/summary.tex
In this section, we present the overall architecture and key components of Matrix-Game 2.0. First, we train a foundation model using our diverse data collection, as detailed in Section \ref{sec:foundtion_model_architecture}. Subsequently, Section \ref{sec:self-forcing} describes our distillation approach that transforms this foundation model into a few-step auto-regressive diffusion model, enabling real-time generation of long video sequences while maintaining visual quality.

%% file: sec/foundation_model.tex
We propose Matrix-Game 2.0, a novel framework to vision-driven world model that explores intelligence capable of understanding and generating the world without relying on language descriptions. In contemporary works, text guidance has become the dominant modality for controlling — examples include SORA\cite{openai2024worldsim}, HunyuanVideo\cite{kong2025hunyuanvideosystematicframeworklarge}, and Wan\cite{wan2025wanopenadvancedlargescale}, all of which leverage text descriptions for generation. However, such methods often introduce semantic priors that bias the generation toward linguistic reasoning rather than physical laws, thereby impeding the model’s ability to grasp the fundamental properties of the visual world.

In contrast, Matrix-Game 2.0 eliminates all forms of language input, focusing solely on learning spatial structures and dynamic patterns from image. This de-semanticized modeling approach is inspired by the concept of spatial intelligence \cite{worldlabs2025generating}, emphasizing that the model’s capabilities should stem from intuitive understanding of visual and physical laws rather than abstract semantic scaffolding.

\begin{figure}[t]
 \centering
 % \hspace*{-0.8cm}
 \includegraphics[width=1\textwidth]{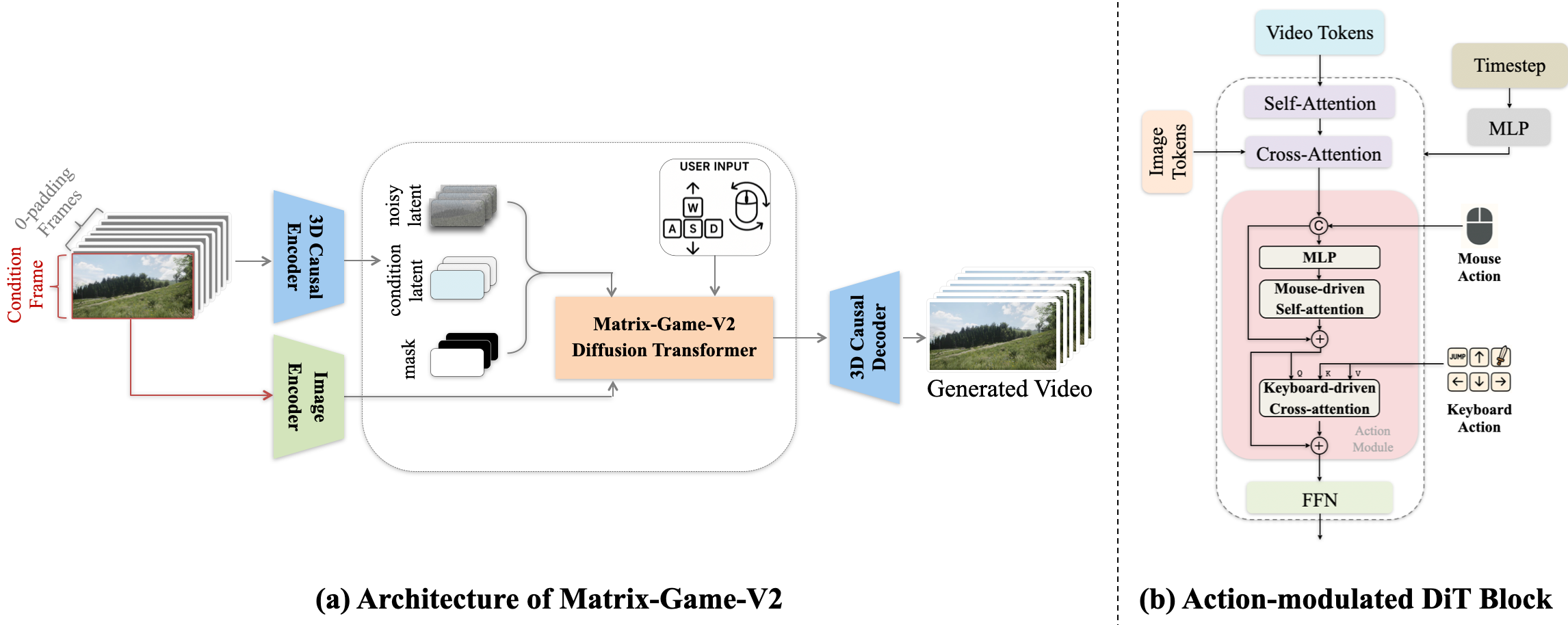}
 \caption{\textbf{Overview of Matrix-Game 2.0 Architecture.} The foundation model is derived from the Wan~\cite{wang2025wan} I2V design. By removing the text branch and adding action modules as in Matrix-Game~\cite{zhang2025matrixgame}, the model predicts next frames only from visual contents and corresponding actions.}
 \label{fig:model_architecture}
 \vspace{-0.3cm}
\end{figure}

As shown in Figure \ref{fig:model_architecture}(a), Matrix-Game 2.0 takes a single reference image and corresponding actions as input, generating a physically plausible video. A 3D Causal VAE~\cite{yu2023language, kingma2013auto} is first employed to compress raw video data along both spatial and temporal dimensions — by a factor of 8$\times$8 in space and 4 in time — enhancing training efficiency and modeling capability. The image input is encoded by 3D VAE encoder and the CLIP image encoder \cite{radford2021learningtransferablevisualmodels} as condition input. Guided by input actions provided by users, the Diffusion Transformer(DiT) generates a visual token sequence, which is subsequently decoded into the video through 3D VAE decoder.

To enable interactions between users and generated contents, Matrix-Game 2.0 incorporates an action module to achieve controllable video generation. Inspired by the control design paradigms of GameFactory~\cite{yu2025gamefactory} and Matrix-Game \cite{zhang2025matrixgame}, we embed frame-level action signals into the DiT blocks, as illustrated in Figure \ref{fig:model_architecture}(b). The injected action signals are divided into two categories: discrete movement actions via keyboard inputs, and continuous viewpoint actions via mouse movements. Specifically, continuous mouse actions are directly concatenated to the input latent representations, forwarded through an MLP layer, and then passed through a temporal self-attention layer. Furthermore, keyboard actions are queried by the fused features through a cross-attention layer, leading to precise controllability for interactions. Different from Matrix-Game~\cite{zhang2025matrixgame}, we use Rotary Positional Encoding~\cite{rope} (RoPE) to replace the sin-cos embeddings added to keyboard inputs to facilitate long video generation.

%% file: sec/self_forcing.tex
Unlike Matrix-Game~\cite{zhang2025matrixgame} which employs a full-sequence diffusion model limited to fixed-length generation, we develop an auto-regressive diffusion model for real-time long video synthesis. Our approach transforms the bidirectional foundation model into an efficient auto-regressive variant through Self-Forcing~\cite{huang2025self}, which addresses exposure bias by conditioning each frame on previously self-generated outputs rather than ground truth. This significantly reduces the error accumulation characteristic of Teacher Forcing~\cite{jin2024pyramidal} or Diffusion Forcing~\cite{chen2024diffusion} approaches.

\begin{figure}[t]
 \centering
 \includegraphics[width=0.98\textwidth]{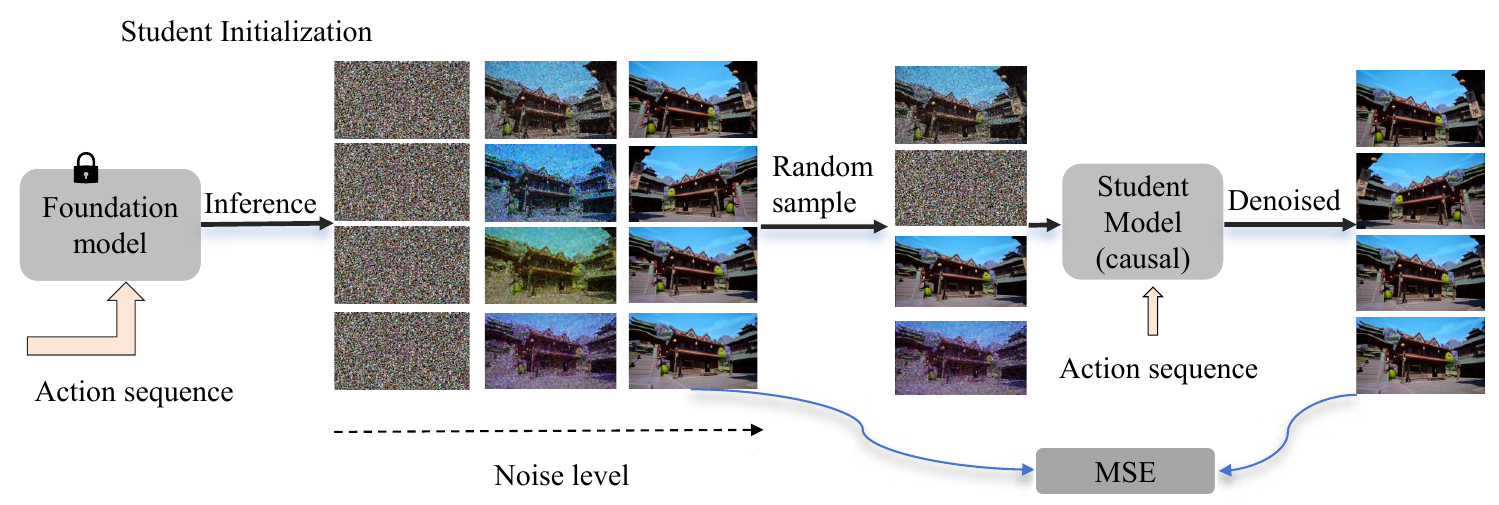}
 \caption{\textbf{Causal Student Model Initialization via ODE Trajectories.} The proposed initialization method stabilizes subsequent distillation training by deriving a few-step causal student model from the bidirectional teacher model through optimal ODE trajectory sampling.}
 \label{fig:ode_init}
\end{figure}

The distillation process comprises two key phases: student initialization and DMD-based~\cite{yin2024one} Self-Forcing training. We first initialize the student generator $G_{\phi}$ with weights from the foundation model, then construct a dataset of ODE trajectories $\left\{ x_t^i \right\}_{i=1}^N$, with $t$ is sampled from 3 steps subset of $[0, T]$.
During training, block-wise causal masks are applied to keys and values in each attention layer. As shown in Figure~\ref{fig:ode_init}, we first sample a sequence of noisy input with $N$ frames from the ODE trajectories and split it into $L$ chunks with independent timesteps $\left\{ x_T^i \right\}_{i=1}^L$. The student generator takes corresponding actions as input and backwards with the regression loss between the denoised output and clean output:
\begin{equation}
\mathcal{L}_{\text{student}} = \mathbb{E}_{x, t^i} \left\| G_{\phi} \left( \left\{ x_{t^i}^i \right\}_{i=1}^L,  \left\{ c^i \right\}_{i=1}^L ,\left\{ t^i \right\}_{i=1}^L \right) - \left\{ x_0^i \right\}_{i=1}^L \right\|^2
\end{equation}

\begin{figure}[t]
 \centering
 \includegraphics[width=1\textwidth]{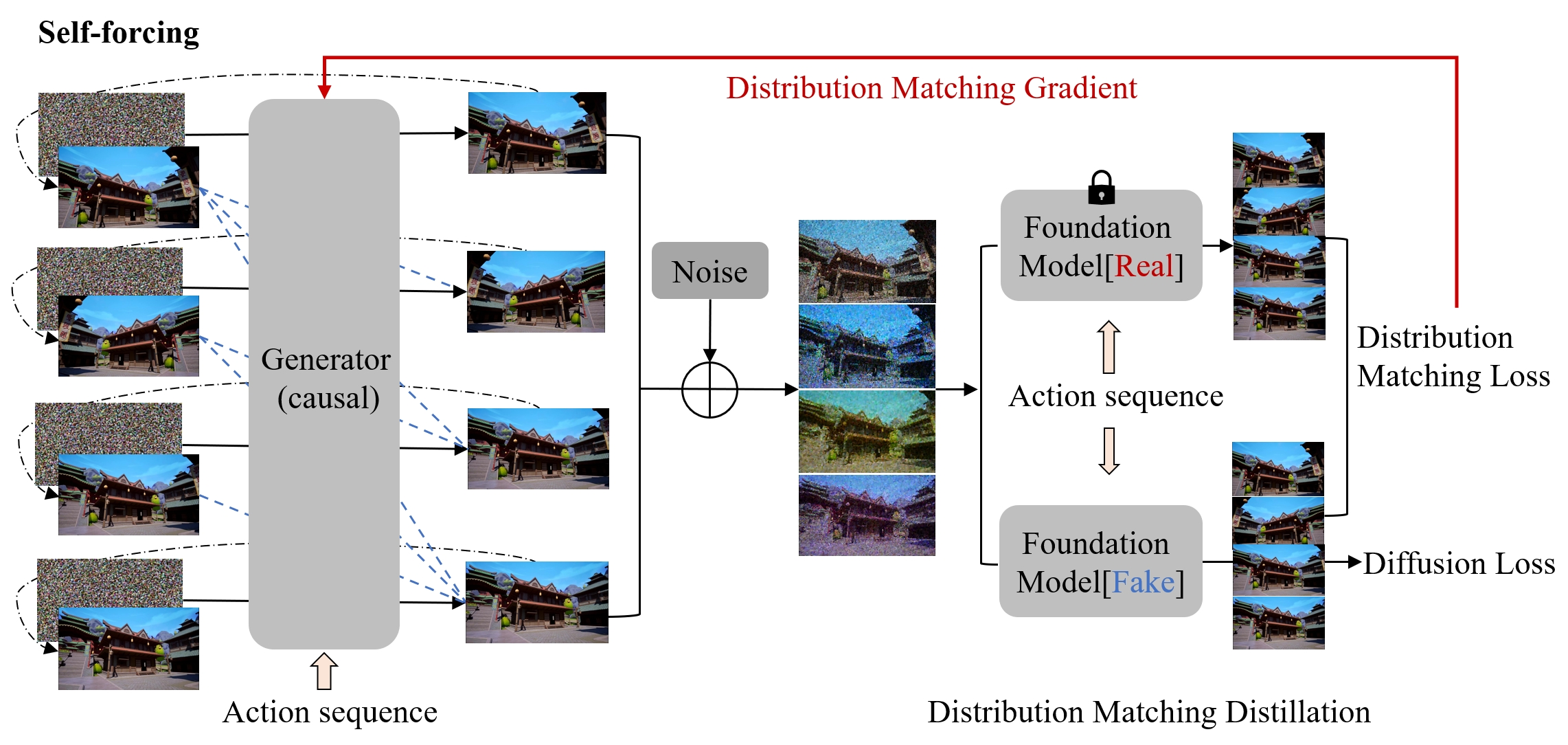}
 \caption{\textbf{Overview of Causal Diffusion Model Training via Self-Forcing.} The distillation process aligns the student model's distributions with the teacher model's through self-conditioned generation. This approach effectively mitigates error accumulation while maintaining the generation quality.}
 \label{fig:dmd}
 \vspace{-0.2cm}
\end{figure}
The subsequent DMD phase (Figure~\ref{fig:dmd}) aligns the student's distributions $p_{\theta, t} \left( x_t^{1:N} \right)$ with the teacher model's $p_{real, t} \left( x_t^{1:N} \right)$ through Self-Forcing. Critically, the generator samples previous frames from its own distribution rather than the ground-truth training data, mitigating training-inference gap and caused error accumulation.

The KV-caching mechanism enables efficient sequential generation by maintaining a fixed-length cache of recent latents and action embeddings. Our rolling cache implementation automatically manages memory by evicting oldest tokens when exceeding capacity, supporting infinite-length generation. To address potential training-inference gap in image-to-video scenarios where the first frame may be excluded during long video inference, we constrain the KV-cache window size. This forces the model to rely more on its learned priors and understanding to the input actions for generation, also improving robustness by making initial frames invisible to subsequent latent frames during training.

%% file: sec/experiments.tex
\begin{figure}[t]
 \centering
 \includegraphics[width=1\textwidth]{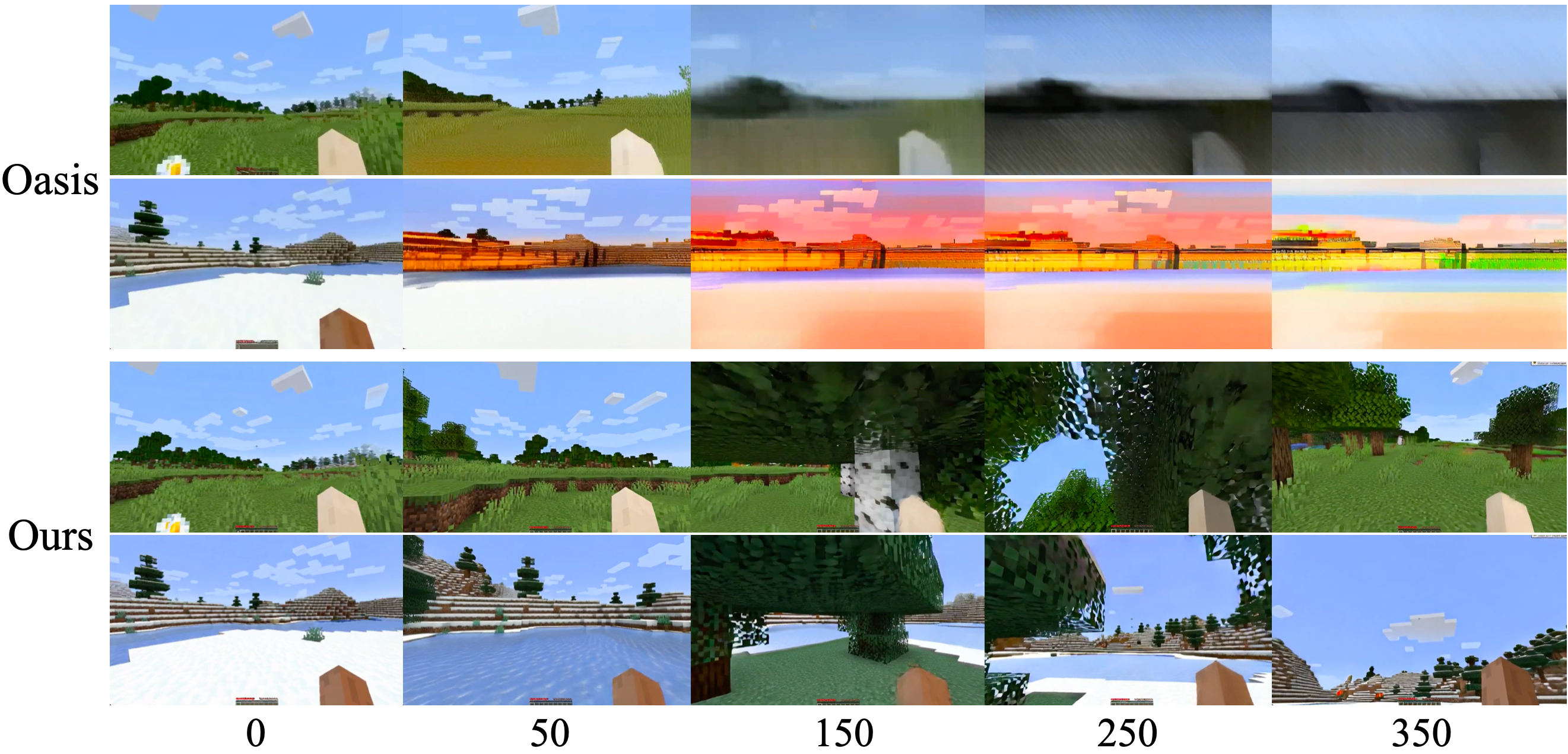}
 \caption{\textbf{Qualitative Comparisons on Minecraft Scene Generations.} Compared to Oasis~\cite{oasis2024}, our model shows superior visual performance in long interactive video generations.}
 \label{fig:mc_comparison}
 \vspace{-0.6cm}
\end{figure}

\begin{table}[t]
% \vspace{-0.2in}
\centering
\caption{\textbf{Quantitative Comparisons on Minecraft Scene Generations.}}
\begin{threeparttable}
\scalebox{0.585}{
\begin{tabular}{l|*{12}{c}}
\toprule
\multirow{2}{*}{Model} & \multicolumn{2}{c}{Visual Quality} && \multicolumn{2}{c}{Temporal Quality} && \multicolumn{2}{c}{Action Controllability} && \multicolumn{2}{c}{Physical Understanding} \\
\cmidrule{2-3}\cmidrule{5-6}\cmidrule{8-9}\cmidrule{11-12}
 & Image Quality $\uparrow$ & Aesthetic $\uparrow$ && Temporal Cons. $\uparrow$ & Motion smooth. $\uparrow$ && Keyboard Acc. $\uparrow$ & Mouse Acc. $\uparrow$ && Obj. Cons. $\uparrow$ & Scenario Cons. $\uparrow$ \\
\midrule
Oasis~\cite{oasis2024} & 0.27 & 0.27 && 0.82 & \textbf{0.99} && 0.73 & 0.56 && 0.18 & \textbf{0.84} \\
Ours & \textbf{0.61} & \textbf{0.50} && \textbf{0.94} & 0.98 && \textbf{0.91} & \textbf{0.95} && \textbf{0.64} & 0.80 \\
\bottomrule
\end{tabular}

}
\end{threeparttable}
\vspace{-0.6cm}
\label{tab:mc_comparison}
\end{table}

\subsection{Experiment Settings}
\paragraph{Implementation Details.}
For training the foundation model, we initialize our model with SkyReels-V2-I2V-1.3B~\cite{chen2025skyreels}, which follows Wan 2.1~\cite{wang2025wan} architecture. The 1.3B variant provides an optimal balance between generation quality and computational efficiency, enabling real-time and high-quality generation performance. We remove the text injection modules from the released checkpoint. To stabilize the whole training process, we firstly fine-tune the model for 5k steps. After that, action modules are added into each DiT block, leading to the total model size as 1.8B. We train the foundation model for 120k steps with learning rate=2e-5, batch size=256.

For distillation, we firstly collect 40k ODE pairs and fine-tune the causal student model for 6k steps, with subsequent 4k training steps via DMD-based Self-Forcing. The learning rate is 6e-6. The chunk size of latent frames and attention local size are set as 3 and 6, respectively. Additionally, self-forcing is a data-free training method, allowing for manual design of the action sequence distribution, which can align better for input actions from users rather than random action sequences produced by automatic scripts.

\paragraph{Dataset.}
The training dataset, produced by the pipeline in Section~\ref{sec:data_pipeline}, consists of about 800-hour action-annotated video at 360p resolution in total. The data includes 153-hour Minecraft video data and 615-hour Unreal Engine data, arranged into 57 frames for each video clip. For real-world scenes, we utilize the open-source Sekai dataset~\cite{li2025sekai}, obtaining an additional 85 hours of training data after data curation. Given that the environment navigation speed and FPS in the Sekai dataset is different from that of Unreal Engine scenes, we perform frame resampling on the Sekai data to align the temporal dynamics and movements. To validate the universality of our framework, we further collect 574-hour game-driver data and 560-hour Temple Run game data, featuring dynamic scenes interaction for additional fine-tuning. All the videos are resized to 352$\times$640 resolution.

\begin{figure}[t]
 \centering
 \includegraphics[width=0.9\textwidth]{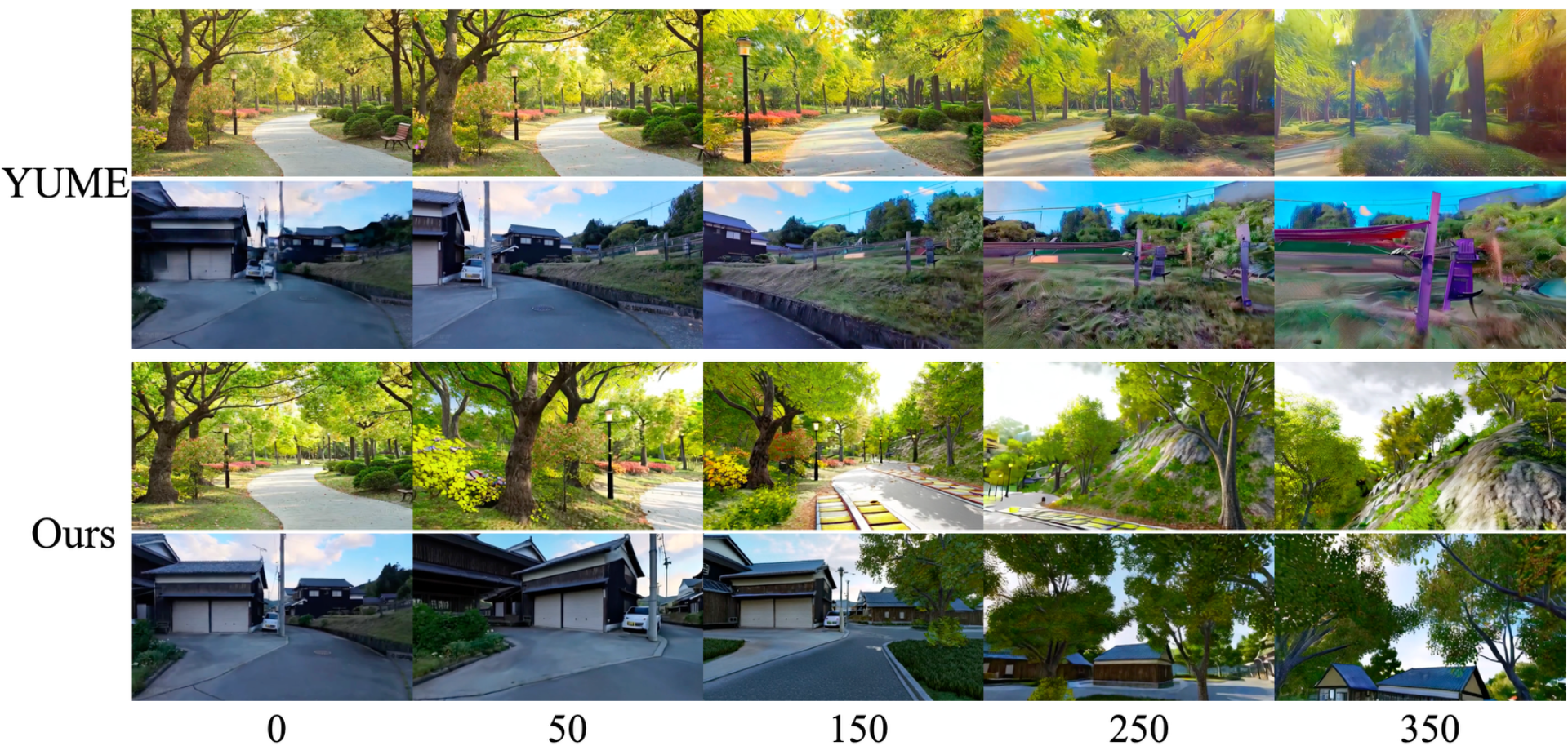}
 \caption{\textbf{Qualitative Comparisons on Wild Scene Generations.} For wild image inputs, Matrix-Game 2.0 exhibits strong generalization capabilities, fast generation speed, and accurate interaction responses.}
 \label{fig:wild_comparison}
 \vspace{-0.4cm}
\end{figure}

\begin{table}[t]
% \vspace{-0.2in}
\centering
\caption{\textbf{Quantitative Comparisons on Wild Scene Generations.}}
\begin{threeparttable}
\scalebox{0.655}{
\begin{tabular}{l|*{12}{c}}
\toprule
\multirow{2}{*}{Model} & \multicolumn{2}{c}{Visual Quality} && \multicolumn{2}{c}{Temporal Quality} && \multicolumn{2}{c}{Physical Understanding} \\
\cmidrule{2-3}\cmidrule{5-6}\cmidrule{8-9}
 & Image Quality $\uparrow$ & Aesthetic $\uparrow$ && Temporal Cons. $\uparrow$ & Motion smooth. $\uparrow$ && Obj. Cons. $\uparrow$ & Scenario Cons. $\uparrow$\\
\midrule
YUME~\cite{mao2025yume} & 0.65 & 0.48 && 0.85 & \textbf{0.99} && \textbf{0.77} & \textbf{0.80} \\
Ours & \textbf{0.67} & \textbf{0.51} && \textbf{0.86} & 0.98 && 0.71 & 0.76 \\
\bottomrule
\end{tabular}
% \vspace{-0.2in}
}
\end{threeparttable}
\vspace{-0.4cm}
\label{tab:wild_comparison}
\end{table}
\paragraph{Evaluation Metrics and Baselines.}
We assess our universal real-time model using the comprehensive GameWorld Score Benchmark~\cite{zhang2025matrixgame} introduced in Matrix-Game 1.0. This benchmark provides a multi-dimensional evaluation framework examining four critical capabilities: visual quality, temporal quality, action controllability and physical rule understanding. Given the current scarcity of open-source interactive world models, we conduct separate evaluations for two distinct domains: Minecraft and wild scenes. For Minecraft environments, we compare against Oasis~\cite{oasis2024} as our primary baseline, while YUME~\cite{mao2025yume} is employed for more complex wild scene generation tasks. All experiments utilize a 597-frame composite action sequence, with evaluation performed on 32 Minecraft scenes and 16 diverse wild scene images, to cover diverse interactive conditions.

\subsection{Generation Results}
We present comprehensive qualitative and quantitative evaluations comparing Matrix-Game 2.0 against state-of-the-art baselines across multiple domains, including long video generation in both Minecraft environments and wild scenes, as well as generation visualization for interactive game driving scenarios and TempleRun game.

\paragraph{Minecraft Scene Results.}
Figure~\ref{fig:mc_comparison} and Table~\ref{tab:mc_comparison} demonstrate Matrix-Game 2.0's superior performance compared to Oasis~\cite{oasis2024}. While Oasis exhibits significant quality degradation after several dozen frames, our model maintains excellent performance throughout extended generation sequences. Quantitative metrics reveal substantial improvements across most evaluation dimensions, though we observe marginally lower scores in scene consistency and action smoothness. We attribute this to Oasis's tendency to produce static frames after collapse, which inflates these particular metrics.

\paragraph{Wild Scene Results.}
Our comparison with YUME~\cite{mao2025yume} in Figure~\ref{fig:wild_comparison} reveals Matrix-Game 2.0's strong robustness in wild scene generation. YUME develops noticeable artifacts and color saturation issues after several hundred frames, while ours maintains stable style fidelity. Moreover, the generation speed of YUME maintains slow, which is hard to be directly applied for interactive world modeling.

\begin{figure}[t]
 \centering
 \includegraphics[width=0.99\textwidth]{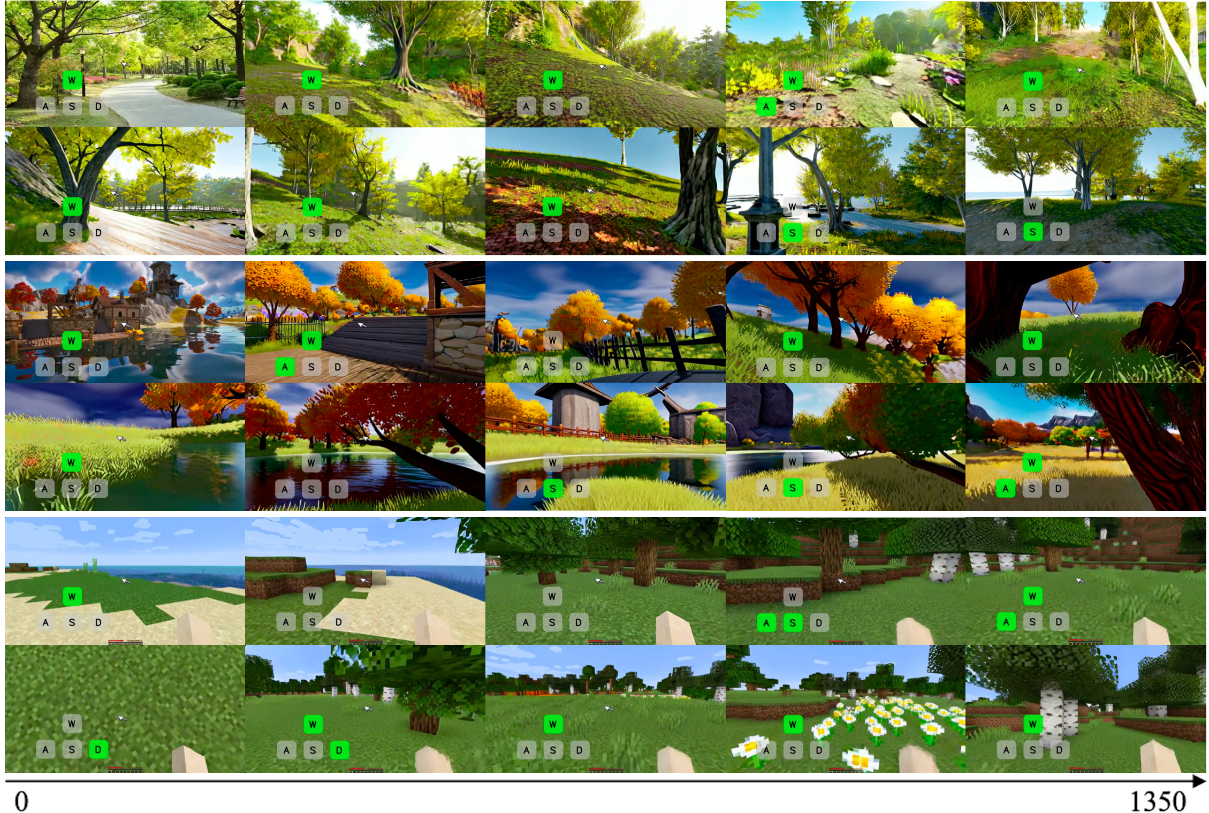}
 \caption{\textbf{Long Video Generations of Matrix-Game 2.0.} The real-time generation results demonstrate excellent visual quality and precise action controllability when generating long videos.}
 \label{fig:long_sample}
 \vspace{-0.4cm}
\end{figure}

Table~\ref{tab:wild_comparison} shows the quantitative results. Since the action controllability assessment in GameWorld Score Benchmark is designed specifically for Minecraft evaluation, it cannot be directly applied to wild scenes. Empirical results demonstrate that YUME exhibits significantly degraded action control performance in out-of-domain scenarios, while our method maintains robust controllability. The generated contents after collapsing of YUME tend to be static, which may also cause higher scores for object consistency and scenario consistency.

\begin{figure}[t]
 \centering
 \includegraphics[width=0.99\textwidth]{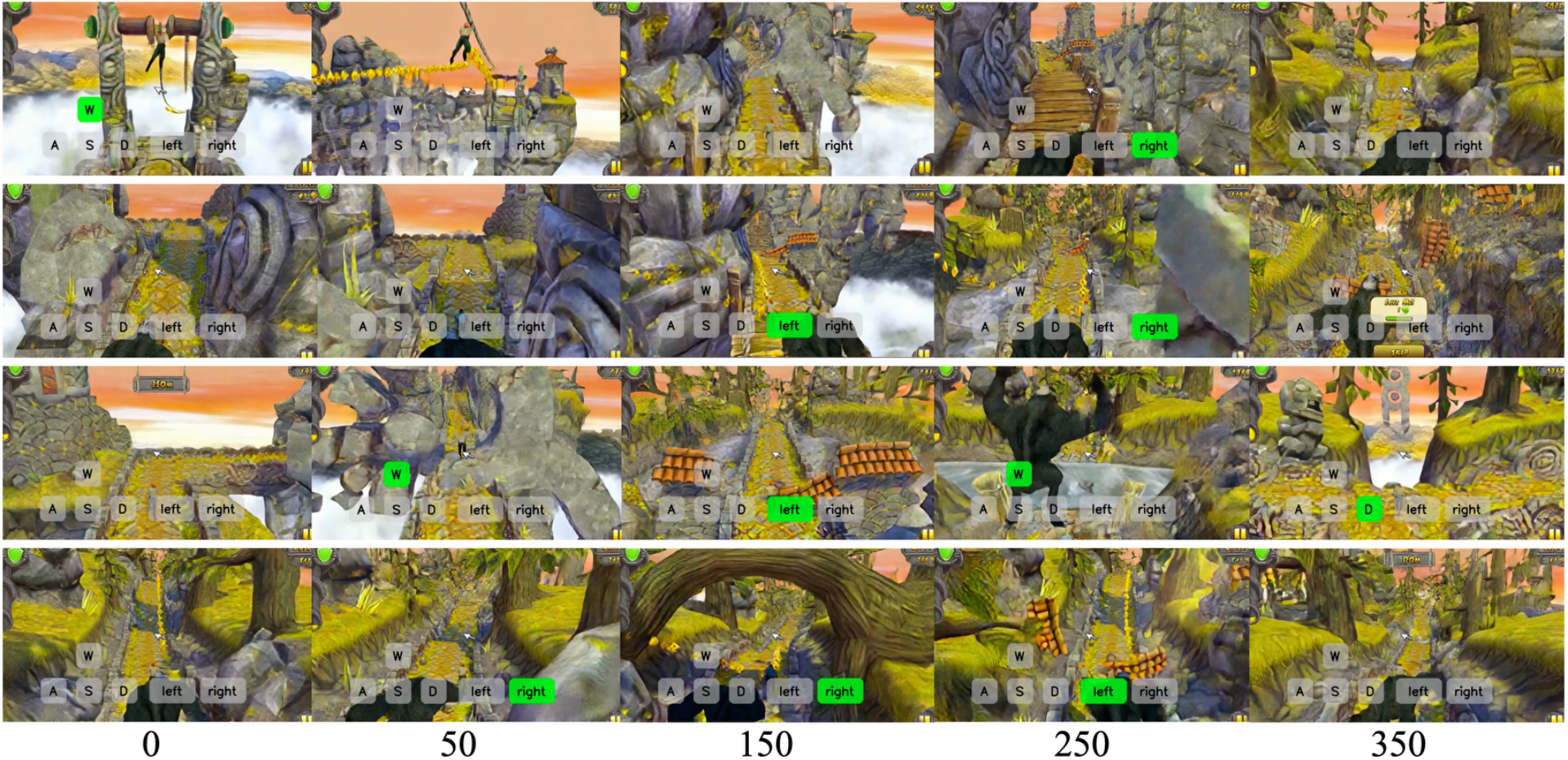}
 \caption{\textbf{Generation Results under the Parkour Game TempleRun scene.}}
 \label{fig:templerun_sample}
 \vspace{-0.4cm}
\end{figure}

\paragraph{More Qualitative Results.}
Figure~\ref{fig:long_sample} showcases Matrix-Game 2.0's exceptional capability for long video generation with minimal quality degradation. The model's strong domain adaptability is further evidenced by its performance in diverse scenarios including interactive driving scenes and TempleRun game (Figure~\ref{fig:templerun_sample}), which also demonstrates its potential as a foundation framework for world modeling.

\begin{figure}[t]
 \centering
 \includegraphics[width=1\textwidth]{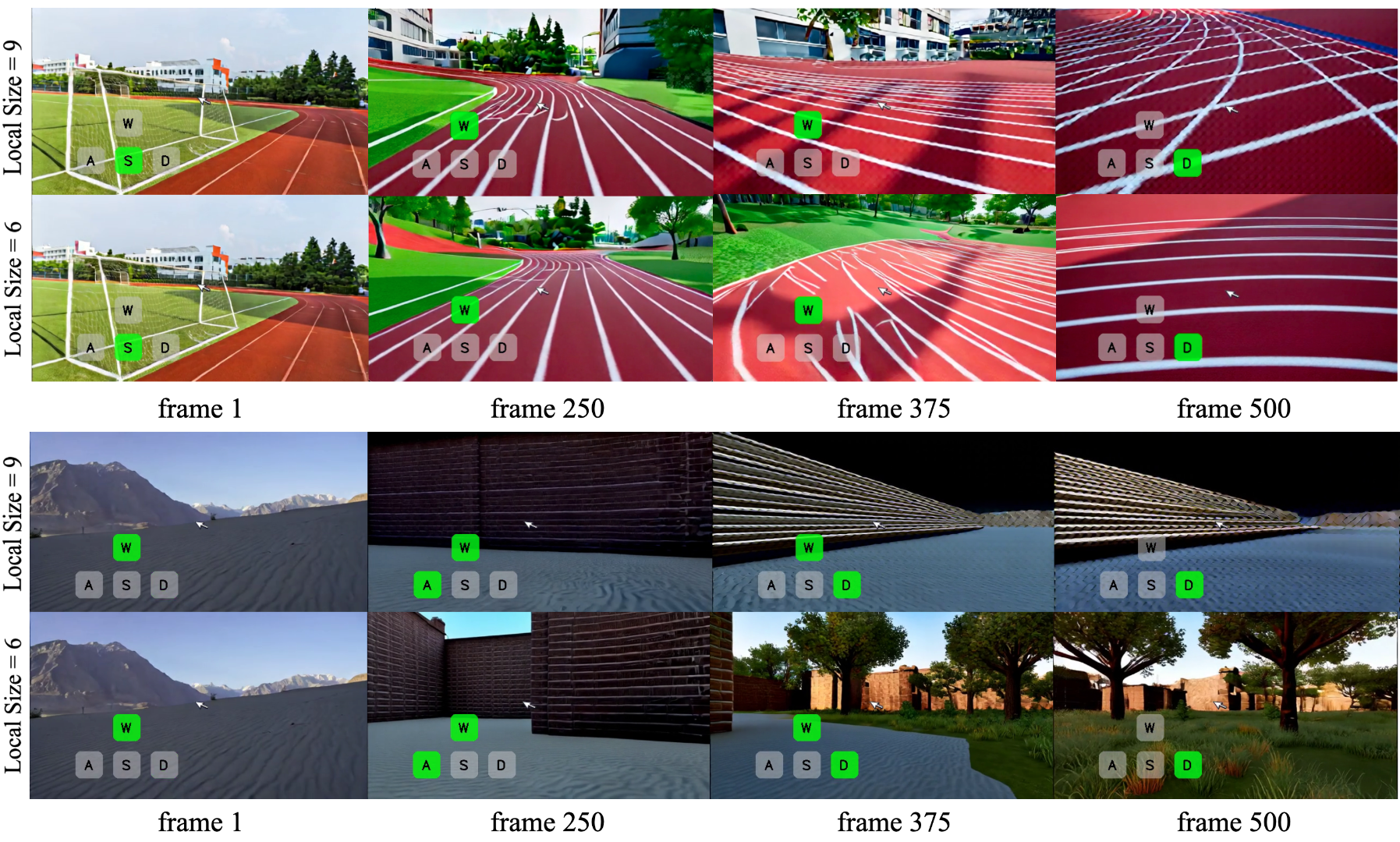}
 \caption{\textbf{Qualitative Comparison on Different Local Size for KV-cache}. Larger local size cause artifacts in long sequences while smaller local size can keep a balance between visual quality and content fidelity.}
 \label{fig:local_size_ablation}
\end{figure}

\subsection{Ablation Studies}
\paragraph{Different KV-cache Local Size.}
The KV-cache mechanism plays a crucial role in maintaining contextual information during Matrix-Game 2.0's auto-regressive generation process. Our investigation reveals an important trade-off in cache size selection: while larger caches (9 latent frames) theoretically provide richer historical context, they paradoxically lead to earlier onset of visual artifacts (Figure~\ref{fig:local_size_ablation}). Comparative analysis shows that models with 6-frame caches demonstrate superior long-term generation quality, with significantly reduced distortion and degradation artifacts.

We attribute this phenomenon to an over-reliance on cached information during generation. With larger cache size, the model increasingly depends on stored cache rather than actively correcting accumulated errors through learned capability of model itself. This creates a compounding effect where artifacts in early frames become more memorized through the cache mechanism, ultimately being treated as valid scene elements. Our empirical study suggests that moderate cache sizes (6 frames) provide a balance between context preservation and error correction capability.

\begin{table}[t]
\centering
\caption{\textbf{Quantitative Comparisons of Different Acceleration Techniques.} While maintaining comparable generation quality metrics, our combined acceleration techniques achieve 25 FPS throughput, enabling on-the-fly video generation.}
\small
\begin{threeparttable}
\scalebox{0.685}{\begin{tabular}{c|*{12}{c}}
\toprule
\multirow{2}{*}{Acceleration Techniques} & \multicolumn{2}{c}{Visual Quality} & \multicolumn{2}{c}{Temporal Quality} & \multicolumn{2}{c}{Action Controllability} & \multicolumn{2}{c}{Physical Understanding} & \multicolumn{1}{c}{Speed} \\
\cmidrule{2-3} \cmidrule{4-5} \cmidrule{6-7} \cmidrule{8-9} \cmidrule{10-10}
 & Image $\uparrow$ & Aesthetic $\uparrow$ & Temporal $\uparrow$ & Motion $\uparrow$ & Keyboard $\uparrow$ & Mouse $\uparrow$ & Object $\uparrow$ & Scenario $\uparrow$ & FPS  $\uparrow$\\
\midrule
(1) +VAE Cache & \textbf{0.61} & \textbf{0.51} & 0.93 & 0.97 & 0.91 & \textbf{0.95} & \textbf{0.68} & \textbf{0.81} & 15.49 \\
(2) (1)+Halving action modules& \textbf{0.61} & \textbf{0.51} & \textbf{0.94} & 0.97 & \textbf{0.92} & \textbf{0.95} & 0.63 & \textbf{0.81} & 21.03 \\
(3) (2)+Reducing denoising steps (4$\rightarrow$3)& \textbf{0.61} & 0.50 & \textbf{0.94} & \textbf{0.98} & 0.91 & \textbf{0.95} & 0.64 & 0.80 & \textbf{25.15} \\
\bottomrule

\end{tabular}}
\vspace{-1.2cm}
\end{threeparttable}
\label{tab:ablation_speed}
\end{table}

\paragraph{Comparative Analysis of Acceleration Techniques.}
To achieve real-time generation at 25 FPS, we systematically optimized both the diffusion model and VAE components through several key modifications. First, we integrated the efficient Wan2.1-VAE architecture with caching mechanism, significantly accelerating the decoding process for extended video sequences. Second, we strategically employ action modules only in the first half of DiT blocks, and reduce the denoising steps from 4 to 3 in the distillation process. The quantitative comparisons are shown in Table~\ref{tab:ablation_speed}. Quantitative comparisons shown in Table~\ref{tab:ablation_speed} demonstrate that these acceleration strategies can achieve 25 FPS while maintaining generation quality, resulting in an optimal speed-quality trade-off.

%% file: sec/conclusion.tex
Matrix-Game 2.0 represents a significant advancement in real-time interactive video generation through carefully constructed data pipeline and effective training framework. First, we developed a comprehensive data production pipeline that overcomes previous limitations in obtaining high-quality training data for interactive scenarios. Our systematic pipeline based on Unreal Engine, together with the video recording framework verified in interactive game environments, establish new standards for scalable production of action-annotated video data at unprecedented fidelity.

Second, we introduced an auto-regressive diffusion framework that combines action-conditioned modulation with distillation based on Self-Forcing. This approach effectively mitigates the error accumulation problem that has traditionally plagued long video synthesis while maintaining real-time performance. Through systematic optimizations of both the diffusion process and VAE architecture, we achieve a generation speed of 25 FPS - for seamless human-in-the-loop interaction.

Extensive experiments demonstrate that Matrix-Game 2.0 sets new benchmarks for interactive generation systems, delivering excellent performance in both visual quality and action controllability. The model's ability to maintain temporal coherence during long-term interactions while responding precisely to user inputs represents a substantial step forward for applications requiring real-time world simulation.
\begin{figure}[t]
 \centering
 \includegraphics[width=0.7\textwidth]{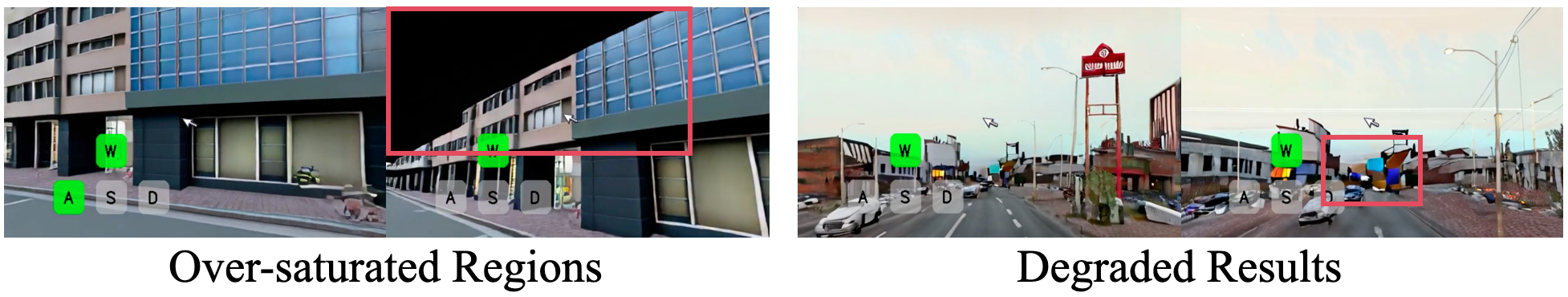}
 \caption{\textbf{Bad cases.} Matrix-Game-V2 sometimes fails when handling out-of-domain scenes, like producing over-saturated (left) or degraded (right) results.}
 \label{fig:limitation}
\end{figure}

\subsection{Limitations}
While demonstrating strong performance, Matrix-Game 2.0 has several limitations that point to future research directions. First, the generalization capability needs to be improved when handling out-of-domain (OOD) scenes - for example, moving the camera upward or step forward for a long time in OOD scenes may result in over-saturated or degraded results. Second, the current 352×640 resolution output falls short of state-of-the-art video generation models that typically produce higher-definition results. Third, while the auto-regressive diffusion model enables long video generation, maintaining content consistency and history over long video generations remains challenging due to the lack of explicit memory mechanisms for history preservation.

We note that these limitations present clear pathways for improvement. The generalization and resolution issues can be improved through expanded training data domain and model architecture scaling. Moreover, the last limitation could be addressed by integrating compatible memory retrieval mechanisms without compromising real-time performance. These directions will be the focus of our future work towards making interactive video generation more practical for real-world applications.